\pdfoutput=1

\documentclass[11pt]{article}

\usepackage{acl}

\usepackage{times}
\usepackage{latexsym}

\usepackage[T1]{fontenc}

\usepackage[utf8]{inputenc}

\usepackage{microtype}

\usepackage{xcolor}
\usepackage{lipsum}

\usepackage{graphicx}

\usepackage{booktabs}

\def\parsum#1{\bgroup \textcolor{blue}{Paragraph summary: #1}\egroup}
\def\sectionsum#1{\bgroup \textcolor{green}{Section content: #1}\egroup \\}

\newcommand{\name}{\texttt{DP-Rewrite}}

\title{DP-Rewrite: Towards Reproducibility and Transparency in Differentially Private Text Rewriting}

\author{Timour Igamberdiev$^{1}$ \and Thomas Arnold$^{2}$ \and Ivan Habernal$^{1}$ \\
	$^{1}$Trustworthy Human Language Technologies \\
	$^{2}$Ubiquitous Knowledge Processing Lab \\
	Department of Computer Science, Technical University of Darmstadt \\
	\texttt{\{timour.igamberdiev,ivan.habernal\}@tu-darmstadt.de}\\
	\texttt{arnold@ukp.informatik.tu-darmstadt.de}\\
	\url{www.trusthlt.org}   $\qquad$\url{www.ukp.tu-darmstadt.de}
}

\begin{document}
    \onecolumn
    \noindent \textbf{DP-Rewrite: Towards Reproducibility and Transparency in Differentially Private Text Rewriting}

    \medskip
    \noindent Timour Igamberdiev, Thomas Arnold and Ivan Habernal

    \bigskip
    This is a \textbf{camera-ready version} of the article accepted for publication at the \emph{29th International Conference on Computational Linguistics (COLING 2022)}. The final official version will be published on the ACL Anthology website later in 2022: \url{https://aclanthology.org/}

    \medskip
    Please cite this pre-print version as follows.
    \medskip

\begin{verbatim}
@InProceedings{Igamberdiev.2022.COLING,
    title = {DP-Rewrite: Towards Reproducibility and Transparency
             in Differentially Private Text Rewriting},
    author = {Igamberdiev, Timour and Arnold, Thomas and
              Habernal, Ivan},
    publisher = {International Committee on Computational
                 Linguistics},
    booktitle = {Proceedings of the 29th International Conference
                 on Computational Linguistics},
    pages = {(to appear)},
    year = {2022},
    address = {Gyeongju, Republic of Korea}
}
\end{verbatim}
    \twocolumn

\maketitle
\begin{abstract}
Text rewriting with differential privacy (DP) provides concrete theoretical guarantees for protecting the privacy of individuals in textual documents. In practice, existing systems may lack the means to validate their privacy-preserving claims, leading to problems of transparency and reproducibility. We introduce \name, an open-source framework for differentially private text rewriting which aims to solve these problems by being modular, extensible, and highly customizable. Our system incorporates a variety of downstream datasets, models, pre-training procedures, and evaluation metrics to provide a flexible way to lead and validate private text rewriting research. To demonstrate our software in practice, we provide a set of experiments as a case study on the ADePT DP text rewriting system, detecting a privacy leak in its pre-training approach. Our system is publicly available, and we hope that it will help the community to make DP text rewriting research more accessible and transparent.
\end{abstract}

\section{Introduction}

Protecting the privacy of individuals has been gaining attention in NLP. One particular setup is text rewriting using local differential privacy (DP) \citep{Dwork.Roth.2013}, which provides probabilistic guarantees of `how much' privacy can be lost in the worst case if an individual gives us their piece of text that has been rewritten with DP.
For instance, given a text ``I want to fly from Newark to Cleveland on Friday'', the system may rewrite it as ``Flights from Los Angeles to Houston this week''.
Only a few recent works have touched on this challenging topic. For example, \citet{krishna2021adept} proposed ADePT: A text rewriting system based on the Laplace mechanism. However, it turned out that their DP method was formally flawed \citep{habernal2021when}. We also see another recent approach, DP-VAE \citep{Weggenmann.et.al.2022.WWW}, which shows results that look surprisingly good for the level of guaranteed privacy. However, neither ADePT nor DP-VAE published their source codes, so the community has no means to perform any empirical checks to validate the privacy-preserving claims. Therefore, the lack of transparency and reproducibility is the main obstacle to the accountability of DP text-rewriting systems.

We asked whether an open, modular, easily extensible, and highly customizable framework for differentially private text rewriting could help the community gain further insight into the utility and potential pitfalls of such systems. We hypothesize that by integrating various downstream datasets, models, pre-training procedures, and evaluation metrics into one software package, we improve the transparency, accountability, and reproducibility of research in differentially private text rewriting.

Our main contributions are twofold. First, this demo paper presents \name, an open-source framework for differentially private text rewriting experiments. It includes a correct reimplementation of ADePT as a baseline, integrates pre-training on several datasets, and allows us to easily perform downstream experiments with varying privacy guarantees by adjusting the privacy budget $\varepsilon$.
Second, \name\ allows us to easily detect another privacy leak in the approach proposed in ADePT, namely in the pre-training strategy of the autoencoder, with the system memorizing the input data.
We demonstrate this in detail as a use case of \name\ in Section \ref{sec:case-study}.\footnote{Our project is available at \texttt{\url{https://github.com/trusthlt/dp-rewrite}}.}

\section{Related Work}

Although the problem of simple data redaction is a widely researched field with several promising approaches \cite{hill2016on, lison2021anonymisation}, the related problem of private text transformation is still largely unexplored.

We only briefly sketch the main ideas of local differentially private algorithms in text rewriting. For a longer introduction to DP see, e.g., \citet{Habernal.2022.ACL,Senge.et.al.2021.arXiv,Igamberdiev.Habernal.2022.LREC}. Let $x, x' \in \mathcal{X}$ be two data points such as texts or vectors, each belonging to a different person. In DP terminology, $x$ and $x'$ are neighboring datasets, as they differ by one person \citep{Desfontaines.Pejo.2020}. A (local) DP algorithm $\mathcal{M}: \mathcal{X} \to \mathcal{Y}$ is a function that takes any single data point $x \in \mathcal{X}$ and produces its `privatized' version $y \in \mathcal{Y}$ which might be an arbitrary object, such as a text or a vector. Privatization is achieved by introducing randomness in $\mathcal{M}$. As a result, $(\varepsilon,0)$-local DP guarantees that for any two neighboring datasets $x, x'$ and any output $y$
\begin{equation}
\ln \left[ 
\frac{\Pr (\mathcal{M}(x) = y)}{\Pr (\mathcal{M}(x') = y)}
\right]
\leq \varepsilon
\end{equation}
where $\varepsilon$ is the privacy budget; the lower, the better privacy is guaranteed. If a text rewriting algorithm satisfies the local DP, it limits the probability of revealing the true text $x$ after observing the privatized text $y$.

\citet{krishna2021adept} proposed ADePT, a DP text rewriting system. It consists of an auto-encoder that learns a compressed latent representation of text, and a DP rewriter that uses the trained auto-encoder, adds Laplace noise to the latent representation vector, and generates the privatized text. Due to a formal error in the scale of the Laplace noise, ADePT violated differential privacy  \citep{habernal2021when}.

\citet{bo2021er} proposed a text rewriting approach that generates words from a latent representation while adding DP noise. However, unlike holistic text rewriting with DP, perturbing text only at the word level does not protect against privacy attacks \citep{Mattern.et.al.2022.arXiv}.

Even more current, \citet{Weggenmann.et.al.2022.WWW} proposed an end-to-end approach to text anonymization using a DP autoencoder, claiming to produce coherent texts of high privacy standards. However, several key aspects of the experiments lack a detailed description, while their results look surprisingly good. Since the code base is not public, we cannot reproduce or reimplement their approach, and we cannot prove or refute our suspicions.

\section{Description of software}

\begin{figure*}[!ht]
    \centering
    \includegraphics[width=\linewidth]{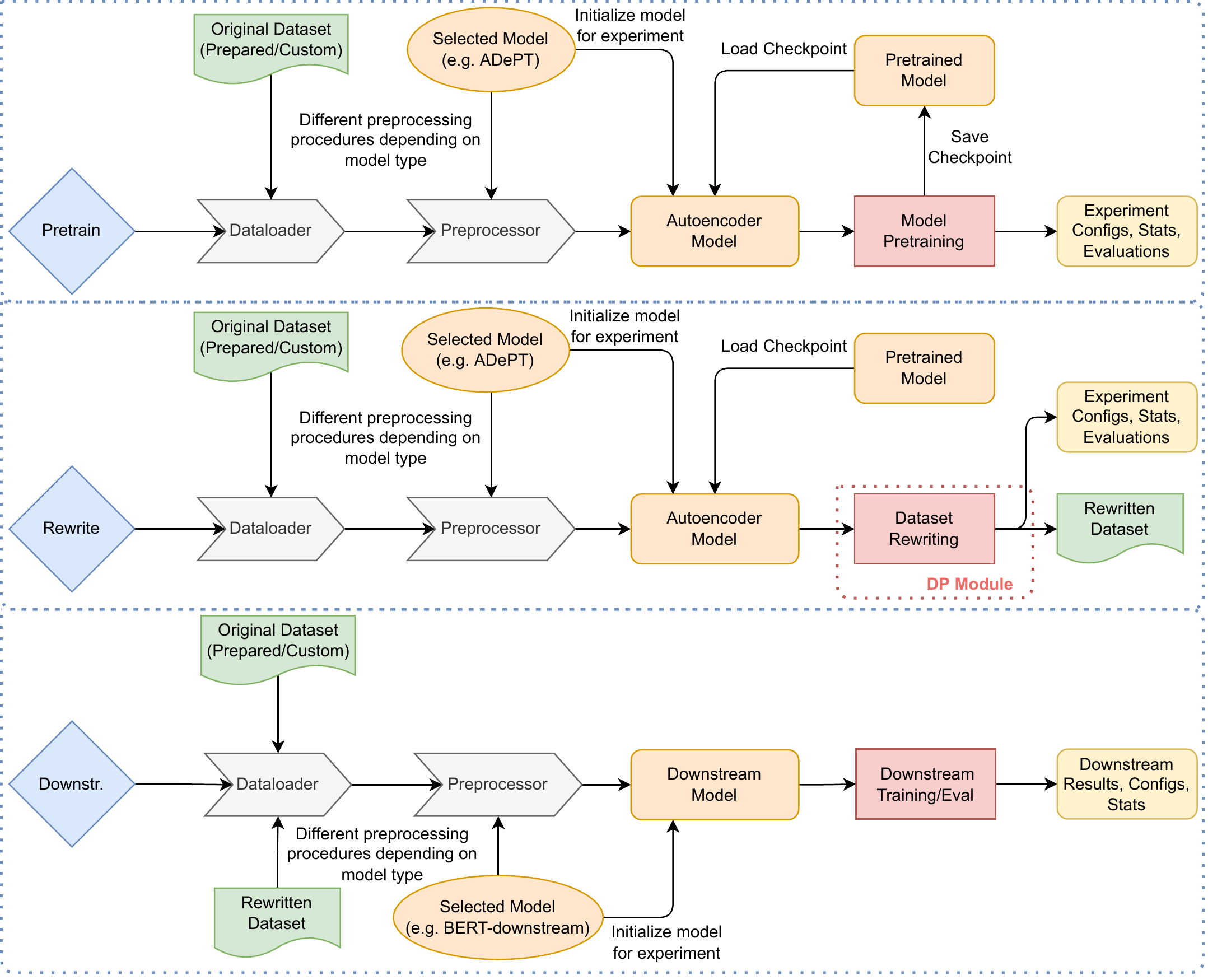}
    \caption{Overall structure of DP-Rewrite. Colors represent groupings of similar components. Blue: Experiment mode. Grey: Dataset preparation. Green: Datasets (original/rewritten). Orange: Model-related components. Red: Main experiment loop. Yellow: Additional experiment outputs.}
    \label{fig:dp-rewrite-diagram}
\end{figure*}

The goal of our system is to provide a seamless way to perform differentially private text rewriting, both on existing and custom datasets. A user can either load a dataset that we provide out-of-the-box, or use a custom one. In addition, we want to make it fast and convenient to run experiments for existing methodologies in DP text rewriting (e.g. ADePT), as well as the ability to integrate novel approaches. For this, we have a general \textit{autoencoder} class based on which out-of-the-box and custom models are built. In this sense, our software is designed to be open and modular, where the researcher can swap out existing components to run a variety of experiments, as well as use the software for one's own privatized text rewriting needs.

The core architecture of our system can be seen in Figure~\ref{fig:dp-rewrite-diagram}. We divide the experiments into three distinct \textbf{modes}, \textit{pre-training}, \textit{rewriting}, and \textit{downstream}. For all three, the pipeline begins with a dataloader which can either be a dataset provided in the framework, or a custom dataset specified by the user. Additionally, a rewritten dataset can be loaded for downstream experiments. The loaded dataset is then preprocessed according to user-specified parameters and the user's selected model, split into different procedures depending on the model type (e.g. RNN-based, transformer-based). The model is then initialized, optionally from an existing checkpoint. At this point, the main experiment is run based on the specified mode, either (1) pre-training the autoencoder, (2) using an existing checkpoint to rewrite the dataset, or (3) running a downstream model on an original or rewritten dataset. For each mode, a variety of evaluations are available, such as BLEU \citep{papineni2002bleu} and BERTScore \citep{zhang2019bertscore} for pre-training and rewriting, and various classification metrics (e.g. $F_1$ score) for downstream experiments. The differential privacy component is incorporated during the rewriting phase for systems such as ADePT, although our framework also allows to incorporate it during the pre-training stage.

\section{Case study}
\label{sec:case-study}

We present here a case study that demonstrates the process of using our framework and provides insights into the ADePT system, for which we provide an implementation in the software.
Our goal is to investigate the difference in rewritten texts and downstream evaluations when we pre-train an autoencoder on one dataset and use this to rewrite another dataset. If we notice a lot of tokens from the dataset used for pre-training in the rewritten dataset,
as well as comparatively higher downstream scores when pre-training and rewriting on the same dataset,
then we can be certain of another form of privacy leakage in ADePT.

\subsection{Datasets}

As in \citet{krishna2021adept}, we use the ATIS \citep{dahl1994expanding} and Snips \citep{coucke2018snips} datasets to conduct experiments on an intent classification task in English.
For both datasets, we use the same train/validation/test split provided by \citet{goo2018slot}, with 4,478 training, 500 validation and 893 test examples for ATIS, and 13,084 training, 700 validation and 700 test examples for Snips. There are a total of 26 intent labels in ATIS and 7 in Snips.

\subsection{Implementation}

We start our experiment pipeline by pre-training two models, one on ATIS (1) and the other on Snips (2), in both cases using the training split.
For pre-training, we set the vocabulary to the maximum number of words from the training set. As in ADePT, we do not incorporate a differential privacy component during pre-training, although we clip encoder hidden representations with a clipping constant value of 5. We limit sequence lengths to a maximum of 20 tokens, pre-training for 200 epochs with a learning rate of 0.003. In contrast to ADePT, we do not use the $\ell_2$ norm for clipping due to issues in privacy guarantees outlined by \citet{habernal2021when} and instead follow the suggested fix for the method by clipping using the $\ell_1$ norm.

We then use these two models for rewriting, applying both pre-trained models for rewriting the training and validation partitions of ATIS and Snips, resulting in four rewriting settings in total. For each setting, we rewrite using five $\varepsilon$ values, $\infty, 1000, 100, 10,$ and $1$. We use the same clipping constant value of 5 as in pre-training.

See Appendix \ref{app:downstream} for details of the downstream experiment setup.

\subsection{Results and analysis}

\begin{figure}[h!]
    \centering
    \includegraphics[width=0.5\textwidth]{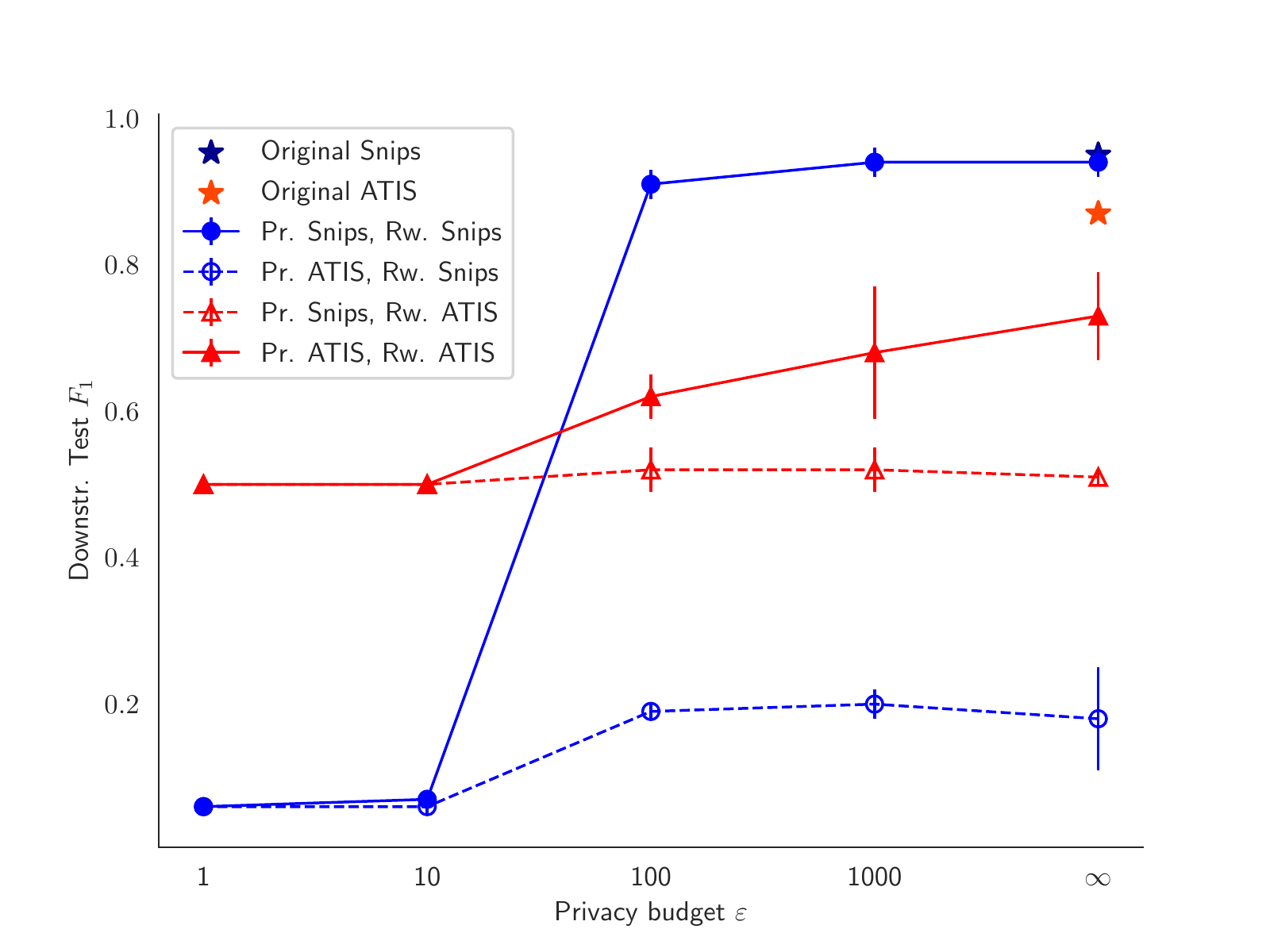}
    \caption{Downstream macro-averaged $F_1$ results for case study experiments with pre-trained and rewritten Snips/ATIS datasets, as well as comparisons with results on the original datasets (``Original Snips'' and ``Original ATIS''). Lower $\varepsilon$ corresponds to better privacy.}
    \label{fig:results-graph}
\end{figure}

\begin{figure}[h!]
    \centering
    \includegraphics[width=\linewidth]{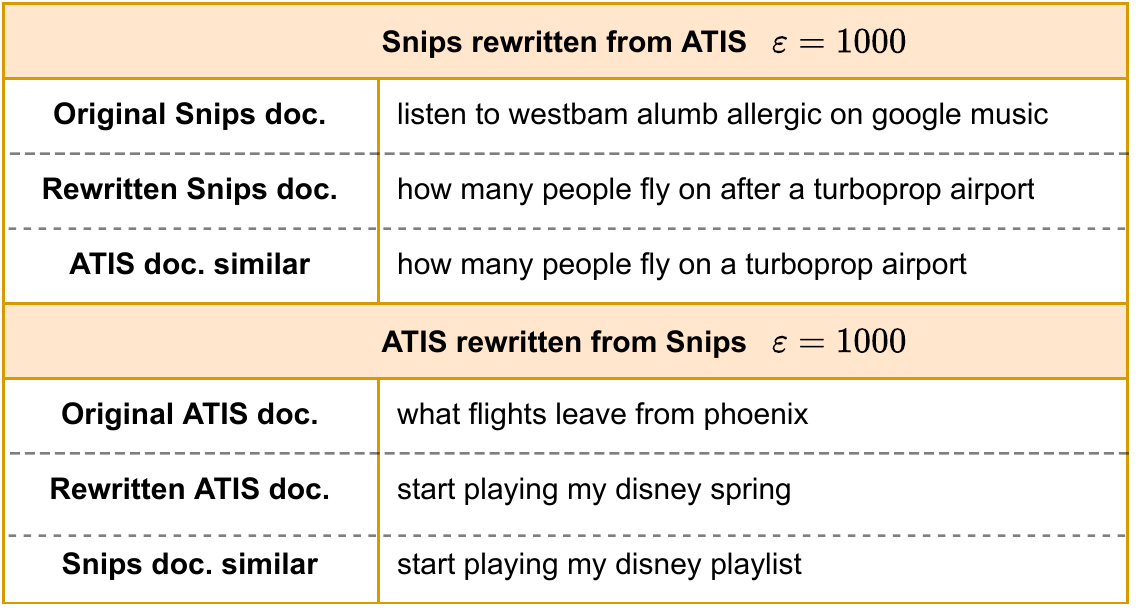}
    \caption{Sample rewritten texts showing memorization by ADePT model when pre-training and rewriting on different datasets. For a given document in the original dataset (``Original Snips/ATIS doc.''), its rewritten version by the model (``Rewritten Snips/ATIS doc.'') has no resemblance to it, but is very similar to another document from the pre-trained dataset (``ATIS/Snips doc. similar'').}
    \label{fig:sample-rewritten}
\end{figure}

Our results can be seen in Figure~\ref{fig:results-graph}. We observe the main patterns as follows. First and most importantly, datasets rewritten using a model that was pre-trained on the same dataset generally show better downstream results than datasets rewritten using a model pre-trained on a different dataset. 
For instance, at $\varepsilon = 1,000$, rewritten Snips from a model pre-trained on Snips has an $F_1$ score of $0.94$, while rewritten Snips from a model pre-trained on ATIS has only $0.20$.
In fact, this is true even at $\varepsilon = \infty$ (non-private setting), without any added noise (e.g. $0.94$ $F_1$ pre-trained Snips, rewritten Snips vs. $0.18$ $F_1$ pre-trained ATIS, rewritten Snips), since for the latter case the model ends up rewriting the dataset that was pre-trained on, having memorized many of its examples.
This can be seen in Figure~\ref{fig:sample-rewritten}, where the rewritten sentences appear to have no resemblance to the original dataset used for rewriting, but are very similar to the data used for pre-training.

Next, as expected, the results decrease for all configurations as the privacy budget $\varepsilon$ decreases, except for rewritten ATIS from a model pre-trained on Snips, where results are low for all $\varepsilon$ values, probably due to the same reasons as shown in Figure~\ref{fig:sample-rewritten}.
At the lower $\varepsilon$ values of $10$ and $1$, performance is very low for all configurations. Since there is so much noise added to the encoder hidden representations, the utility of ADePT's rewriting is severely diminished, for any data inputs.

Finally, compared to running downstream experiments on the original dataset,
Snips rewritten with a model pre-trained on Snips shows about the same results at high $\varepsilon$ values (e.g. $0.94$ $F_1$ pre-trained Snips, rewritten Snips vs. $0.95$ $F_1$ original Snips for $\varepsilon=\infty$). ATIS rewritten with a model pre-trained on ATIS shows lower results in this case (e.g. $0.73$ $F_1$ pre-trained ATIS, rewritten ATIS vs. $0.87$ $F_1$ original ATIS for $\varepsilon=\infty$).
We speculate that since ATIS is a smaller dataset, there are less data points to effectively pre-train ADePT for the autoencoding task. We additionally report random and majority baselines in Appendix \ref{app:table} on the original datasets for comparison.

We have thus shown that, despite fixing the theoretical privacy guarantees of ADePT, the pre-training procedure still results in privacy leakage, with rewritten datasets exposing a lot of information from the dataset used for pre-training. As a result, downstream performance is inflated if the datasets for pre-training and rewriting are the same.

\section{Conclusion}

We introduced \name, an open-source framework for differentially private text rewriting experiments.
We have demonstrated a sample use-case for our framework, which allows us to detect privacy leakage in the pre-training procedure of the ADePT system,
an example of how the modular and customizable nature of the software allows for transparency and reproducibility in DP text-rewriting research.
\name\ is continuing to be under active development, and we are incorporating new datasets and private text rewriting methodologies as they are released. We welcome feedback from the community.

\section*{Acknowledgements}

The independent research group TrustHLT is supported by the Hessian Ministry of Higher Education, Research, Science and the Arts. This project was partly supported by the National Research Center for Applied Cybersecurity ATHENE. Thanks to Johannes Gaese and Jonas Rikowski who contributed in the initial implementation phases.

\bibliography{bibliography}
\bibliographystyle{acl_natbib}

\clearpage

\appendix

\section{Detailed results of the case study}
\label{app:table}

\begin{table}[h!]
	\centering
	\begin{tabular}{llr|l}
		\textbf{Pretr. Dat.}	& \textbf{Rewr. Dat.}	& \textbf{$\varepsilon$}	&\textbf{Test $F_1$}		\\ \midrule
		Snips	& Snips	& $\infty$	& 0.94 (0.02)	\\
		Snips	& Snips	& 1,000	& 0.94 (0.02)		\\
		Snips	& Snips	& 100	& 0.91 (0.02)	\\
		Snips	& Snips & 10 	& 0.07 (0.01) 	\\
		Snips	& Snips	& 1	& 0.06 (0.00)	\\
		ATIS	& Snips	& $\infty$	& 0.18 (0.07)	\\
		ATIS	& Snips	& 1,000	& 0.20 (0.02)		\\
		ATIS	& Snips	& 100	& 0.19 (0.01)	\\
		ATIS	& Snips & 10 	& 0.06 (0.01) 	\\
		ATIS	& Snips	& 1	& 0.06 (0.01)	\\
		Snips	& ATIS	& $\infty$	& 0.51 (0.01)	\\
		Snips	& ATIS	& 1,000	& 0.52 (0.03)		\\
		Snips	& ATIS	& 100	& 0.52 (0.03)	\\
		Snips	& ATIS & 10 	& 0.50 (0.01) 	\\
		Snips	& ATIS	& 1	& 0.50 (0.01)	\\
		ATIS	& ATIS	& $\infty$	& 0.73 (0.06)	\\
		ATIS	& ATIS	& 1,000	& 0.68 (0.09)		\\
		ATIS	& ATIS	& 100	& 0.62 (0.03)	\\
		ATIS	& ATIS & 10 	& 0.50 (0.01) 	\\
		ATIS	& ATIS	& 1	& 0.50 (0.01)	\\
		\midrule
		\multicolumn{3}{l|}{\textbf{Snips Orig.}}	& 0.95 (0.01) \\
		\multicolumn{3}{l|}{\textbf{ATIS Orig.}}	& 0.87 (0.03) \\
		\multicolumn{3}{l|}{\textbf{Snips Rand.}}	& 0.14 \\
		\multicolumn{3}{l|}{\textbf{ATIS Rand.}}	& 0.01 \\
		\multicolumn{3}{l|}{\textbf{Snips Maj.}}	& 0.03 \\
		\multicolumn{3}{l|}{\textbf{ATIS Maj.}}	& 0.13 \\
	\end{tabular}
	\caption{\label{tab1:results} Downstream macro-averaged $F_1$ results for case study experiments with pre-trained and rewritten Snips/ATIS datasets. We additionally show results on the original datasets, as well as random and majority baselines. Test $F_1$ shown as ``mean (standard deviation)'' over five runs with different random seeds. Lower $\varepsilon$ corresponds to better privacy.}
\end{table}

\section{Downstream experiment setup}
\label{app:downstream}

For downstream experiments, we use a pre-trained BERT model \citep{devlin2018bert}, with an additional feedforward layer that takes the mean of the last hidden states as input and predicts the output label. We use the rewritten training and validation sets for each configuration, and the original test sets for final evaluation. We run each configuration with five different random seeds and report the mean and standard deviation.

\end{document}